\PassOptionsToPackage{table}{xcolor}
\documentclass[dvipsnames,sigconf=true, nonacm=true, review=false, anonymous = false,]{acmart}
\usepackage{booktabs} 

\copyrightyear{2024}
\acmYear{2024}

\copyrightyear{2024}
\acmYear{2024}
\setcopyright{rightsretained}
\acmConference[GECCO '25 Companion]{Genetic and Evolutionary Computation Conference}{July 14--18, 2025}{Malaga, Spain}
\acmBooktitle{Genetic and Evolutionary Computation Conference (GECCO '25), July 14--18, 2025, Malaga, Spain}

\acmPrice{15.00}
\acmDOI{10.1145/3638530.3664124}
\acmISBN{979-8-4007-0495-6/24/07}

\usepackage{amsmath}
\usepackage{xspace}
\usepackage{xcolor}
\usepackage{dsfont}
\usepackage{rotating}
\usepackage{makecell}
\usepackage{multirow}
\usepackage{subcaption}
\usepackage{placeins,booktabs}

\usepackage{algorithm}
\usepackage{algpseudocode}
\usepackage{float}

\usepackage{amsthm}
\usepackage{amsmath}
\renewcommand{\epsilon}{\varepsilon}



\begin{document}
\title[Code Evolution Graphs]{Code Evolution Graphs: Understanding Large Language Model Driven Design of Algorithms}



\author{Niki van Stein}
\email{n.van.stein@liacs.leidenuniv.nl}
\orcid{0000-0002-0013-7969}
\affiliation{%
  \institution{LIACS, Leiden University}
  \streetaddress{Einsteinweg 55}
  \city{Leiden}
  \country{Netherlands}
  \postcode{2333 CC}
}

\author{Anna V. Kononova}
\email{a.kononova@liacs.leidenuniv.nl}
\orcid{0000-0002-4138-7024}
\affiliation{%
  \institution{LIACS, Leiden University}
  \streetaddress{Einsteinweg 55}
  \city{Leiden}
  \country{Netherlands}
  \postcode{2333 CC}
}

\author{Lars Kotthoff}
\email{larsko@uwyo.edu}
\orcid{0000-0003-4635-6873}
\affiliation{%
  \institution{University of Wyoming}
  \streetaddress{1000 E University Ave}
  \city{Laramie}
  \country{United States}
  \postcode{WY 82071-2000}
}
\author{Thomas B{\"a}ck}
\email{t.h.w.baeck@liacs.leidenuniv.nl}
\orcid{0000-0001-6768-1478}
\affiliation{%
  \institution{LIACS, Leiden University}
  \streetaddress{Einsteinweg 55}
  \city{Leiden}
  \country{Netherlands}
  \postcode{2333 CC}
}

\renewcommand{\shortauthors}{N.\ van Stein et al.}

\begin{abstract}
Large Language Models (LLMs) have demonstrated great promise in generating code, especially when used inside an evolutionary computation framework to iteratively optimize the generated algorithms. However, in some cases they fail to generate competitive algorithms or the code optimization stalls, and we are left with no recourse because of a lack of understanding of the generation process and generated codes. We present a novel approach to mitigate this problem by enabling users to analyze the generated codes inside the evolutionary process and how they evolve over repeated prompting of the LLM. We show results for three benchmark problem classes and demonstrate novel insights. In particular, LLMs tend to generate more complex code with repeated prompting, but additional complexity can hurt algorithmic performance in some cases. Different LLMs have different coding ``styles'' and generated code tends to be dissimilar to other LLMs. These two findings suggest that using different LLMs inside the code evolution frameworks might produce higher performing code than using only one LLM.
\end{abstract}

\maketitle

\section{Introduction}
The automated generation and optimization of algorithms has become a critical area of research in evolutionary computation and artificial intelligence. Traditional approaches to designing optimization algorithms often rely on expert knowledge, which can be time-consuming and prone to human bias. Recently, Large Language Models (LLMs) such as GPT-4 \cite{openai2023chatgpt4o} have demonstrated significant potential for automating this process by synthesizing and refining algorithms based on natural language prompts and feedback \cite{vanstein2024llamealargelanguagemodel,zhang2024understanding,fei2024eoh,FunSearch2024}.

One of the key challenges in automated algorithm design with LLMs is the limited direct control we have over the algorithm generation, mutation and crossover. This leads to a limited understanding of how the evolutionary search through code space is performing and how we can improve these algorithms in a systematic way. To address this, we propose the concept of \textit{Code Evolution Graphs} (CEGs), which integrate static code analysis, graph-based representations, and complexity metrics to provide insights into optimization dynamics and the resulting algorithmic behaviors of code generation and optimization frameworks. By utilizing Abstract Syntax Trees (ASTs) as a basis, CEGs enable the extraction of static code features, offering a comprehensive toolbox for understanding the evolution of algorithms over successive generations.

In this paper, we focus on analyzing the structural and complexity characteristics of algorithms generated by the Large Language Model Evolutionary Algorithm (LLaMEA) \cite{vanstein2024llamealargelanguagemodel}, LLaMEA with in-the-loop hyper-parameter optimization (LLaMEA-HPO) \cite{vanstein2024intheloophyperparameteroptimizationllmbased} and Evolution of Heuristics (EoH) \cite{fei2024eoh}, across three benchmark tasks: Black-Box Optimization (BBO), Online Bin Packing (OBP), and the Traveling Salesperson Problem (TSP). We aim to answer key questions about the nature of the algorithms produced, such as their structural modularity, logical complexity, and scalability and how these features evolve during the (meta)optimization runs.


The main contributions of this work are as follows:
\begin{itemize}
    \item We introduce Code Evolution Graphs, a novel methodology that combines graph-based representations and complexity analysis to evaluate the evolution of code generated by LLMs and evolutionary frameworks.
    \item We systematically extract and analyze 20 structural and complexity features from ASTs, including cyclomatic complexity, token counts, clustering coefficients, and entropy measures, to uncover patterns in algorithmic design.
    \item We provide a comparative analysis of LLaMEA and EoH across diverse optimization tasks, revealing the strengths and limitations of LLM-driven algorithm generation.
    \item We visualize the evolution of code with PCA and t-SNE to uncover high-level trends and structural shifts that provide novel insights into the optimization process.
\end{itemize}

Recent advancements in LLM-based search frameworks, such as Funsearch~\cite{FunSearch2024}, LLaMEA~\cite{vanstein2024llamealargelanguagemodel} and EoH~\cite{fei2024eoh}, have demonstrated the feasibility of using natural language processing models to generate novel algorithms whose performance rivals that of state-of-the-art heuristic methods. Funsearch was among the first methods that showed that with evolutionary search techniques LLMs could generate novel and well performing heuristics; however, Funsearch required a few million evaluations to do so at that time.
LLaMEA leverages an iterative process of algorithm generation, evaluation, and refinement, guided by fitness feedback and error correction, to evolve complex optimization strategies, originally proposed to solve black-box optimization tasks \cite{vanstein2024llamealargelanguagemodel}. In contrast, EoH focuses on the evolution of smaller heuristic functions through mutation and crossover operations \cite{fei2024eoh}, originally aimed at solving combinatorial problems. Both approaches have shown promising results on benchmark tasks, but we lack a detailed understanding of the structural and functional properties of the generated code and the evolution thereof.

The introduction of Code Evolution Graphs addresses this gap by providing a unified framework for analyzing a wide range of attributes of the codes generated over the optimization run. By leveraging AST-based feature extraction and graph-theoretic metrics, we aim to gain deeper insights into the design principles underlying LLM-generated algorithms and their impact on performance.




\section{Related Work}
The automated generation and optimization of algorithms has been a topic of discussion in the optimization community since at least the early 2010s, when \cite{hoos_programming_2012} introduced the concept of \textit{Programming by Optimization}. To avoid premature commitment to design choices in the algorithm implementations, developers were encouraged instead to maintain a selection of alternatives, forming a parameterized combinatorial algorithm design space that can be searched by a meta-optimizer for better empirical performance over representative problems. The concept has been subsequently practically extended to allow continuous/mixed design spaces~\cite{vanRijn2016} and introduced to numerical black-box optimization context as \textit{modular algorithms}~\cite{Vermetten2019}, where a meta-optimizer searches over algorithms with a defined structure and a known range of options for all mo\-du\-les. These developments have subsequently been recognized as part of the broader task of \textit{automated algorithm design} (AAD) for optimisation~\cite{vermetten_DagRep.13.8.46}. With the advent of LLM-based approaches, we argue here for the extension of AAD to include the automated generation and optimization of algorithms without prior specification of algorithm structure. Thus, in the remainder of this paper, we focus exclusively on AAD for the generation and optimization of meta-heuristic for optimization tasks, referring to them as AAD.

Analyzing the behavior of optimization algorithms has been a central theme in black-box optimization research, with various approaches proposed to gain insights into algorithm performance, search dynamics, and problem landscapes. While this paper represents the first effort to analyze code evolution in the AAD 
domain, we draw upon related work in algorithm trajectory analysis, landscape modeling, and static code analysis to ground our approach. 

\subsection{Analysis of Optimization Algorithm Behavior}

Search Trajectory Networks (STNs) have been employed to visualize and understand the movement of algorithms in the search space \cite{OCHOA2021107492}. STNs represent the trajectory of an algorithm as a directed graph, where nodes correspond to search locations and edges denote transitions between them. This approach has been particularly useful for identifying common pathways and termination points in the search process. 
However, STNs typically do not capture temporal information, such as how long an algorithm remains in a particular region. Attractor networks, introduced in \cite{thomson2024stallingspaceattractoranalysis}, address this limitation by focusing on regions of stagnation, or ``attractors'' where an algorithm stalls for a predefined number of evaluations.
These attractor networks provide a coarser-grained view of algorithm dynamics, revealing insights into stalling behavior that are not evident in traditional STNs.

Another form of attractor analysis for continuous optimization, is by using local optima networks (LONs) \cite{ochoa2014local}. LONs represent the connectivity and distribution of local optima in a search space, providing insights into the basins of attraction. However, LONs often rely on local search methods for their construction, limiting their generalization to other algorithms. 

Exploratory Landscape Analysis (ELA) provides another lens for understanding optimization problems by characterizing the fitness landscape using features such as ruggedness, separability, and global structure \cite{ela_mersmann2011}. These features are essential for predicting algorithm performance and tailoring strategies to specific problem characteristics. While ELA focuses on the BBO problem landscape, our method focuses on the AAD problem landscape, and more specifically the structural properties of the algorithms found during the search, extracting features from the code to understand how they evolve and adapt to various tasks.

While the above methods are applied to gain insights into algo\-rithm-problem interactions, they cannot be (straightforwardly) applied to AAD tasks. The main challenge is that ``individuals'' in the AAD task are algorithms, with no clearly defined representation that can be used to compute similarities. While some work represents algorithms as vectors based on pre-defined modules of the algorithms~\cite{vanstein2024explainable}, this approach is not possible if the code is generated and does not adhere to any pre-defined modular structure.

\subsection{Static Code Analysis}

Static code analysis, a well-established technique in software engineering, has been instrumental in assessing the complexity, maintainability, and quality of source code \cite{nunez2017source}. Metrics such as cyclomatic complexity and code token analysis provide quantitative measures of code structure. When we speak of `tokens' in this work, we mean lexical tokens of the code, indicating total code size.

The work by Pulatov et al.~\cite{pulatov2022opening} introduces an approach to improving algorithm selection by analyzing structural features of algorithms, moving beyond traditional ``black-box'' methods that rely solely on performance observations. By using static code analysis and AST features such as cyclomatic complexity and clustering coefficients, the authors demonstrate that incorporating algorithm features into the selection process can lead to performance improvements. This perspective aligns with our goal of understanding the evolution of LLM-generated algorithms by analyzing their structural and complexity characteristics, as captured by AST-based metrics.
Inspired by this, our work applies static code analysis to evaluate the structural and complexity features of algorithms generated in the AAD domain to provide a comprehensive understanding of code evolution.

Visualization techniques for changes in code over time were proposed in the past to mainly visualize changes in repositories or software libraries. In \cite{telea2008code} for example, the authors introduce the Code Flows methodology, a visualization technique for understanding structural changes in source code across multiple versions. 
Instead of visualizing a complete repository of code based on a Git or change log, we propose a visualization and analysis methodology for evolving code (single-file algorithms or even single functions) generated by LLMs in an AAD setting.


Despite these advancements, no prior work has systematically analyzed the structural and complexity characteristics of algorithms generated in the AAD domain. This gap motivates our study, which combines techniques from STNs, attractor analysis, and static code analysis to offer a novel perspective on algorithm design and evolution. By extracting and analyzing a diverse set of features from LLM-generated code, we aim to uncover patterns and principles that underlie successful algorithm generation, aiming to contribute to a deeper understanding of the AAD landscape.

\section{Methodology}\label{sec:methodology}
In this section, we introduce the methodological framework for analyzing the evolution of code generated during the optimization process using Large Language Models and evolutionary frameworks. Our approach focuses on extracting meaningful insights into the properties of the generated algorithms that affect performance. To achieve this, we leverage Abstract Syntax Trees as a foundational representation of the code, from which we derive a comprehensive set of metrics and features. These include structural properties, complexity measures, and graph-based representations that collectively characterize the behavior of the algorithms.

We begin by describing the \textit{Abstract Syntax Tree Features} (Section~\ref{sec:ast_features}), which capture the syntactic structure of the code and enable the computation of graph-theoretic metrics. Next, we introduce the \textit{Code Complexity Features} (Section~\ref{sec:code_complexity}) we use, focusing on cyclomatic complexity, token counts, and parameter counts to quantify the logical and structural complexity of the generated algorithms. Finally, we formally define the concept of \textit{Code Evolution Graphs} (Section~\ref{sec:code_evolution_graphs}), a novel graph-based representation that models the lineage and transformations of code over successive generations in the optimization process.

\subsection{Abstract Syntax Tree Features}
\label{sec:ast_features}
Abstract Syntax Trees serve as a foundational representation of source code structure, enabling the extraction of syntactic features for code analysis. ASTs can be seen as directed acyclic graphs (DAGs), where nodes represent syntactic elements (e.g., loops, conditionals), and edges capture their hierarchical relationships. 
This facilitates the computation of graph-theoretic metrics and other structural properties.

To characterize ASTs, we extract the following features:
\begin{description}
    \item[Structural Properties] Metrics such as node count, edge count, and depth distribution quantify code complexity.
    \item[Graph Centrality] Eigenvector centrality highlights critical nodes influencing graph connectivity.
    \item[Clustering Coefficients] Measures of node clustering reveal the modularity of the code structure.
    \item[Transitivity and Assortativity] These metrics reflect code cohesiveness and structural correlations.
    \item[Entropy Measures] Degree and depth entropy quantify the diversity and balance of the AST structure.
\end{description}

This comprehensive set of metrics enables a comparative analysis of algorithmic structures and diversity and serves as a basis for visualizing code evolution.

\subsection{Code Complexity Features}
\label{sec:code_complexity}
Understanding the complexity of generated code is essential for evaluating its scalability, maintainability, and computational efficiency. To this end, we leverage a range of static analysis metrics to quantify the complexity of code structures produced during the optimization process. These metrics are particularly important for assessing the trade-offs between algorithmic sophistication and runtime performance.

The following key complexity features are extracted:
\begin{description}
    \item[Cyclomatic Complexity] This metric quantifies the number of linearly independent paths through the code, providing a measure of its logical complexity~\cite{ebert2016cyclomatic}.
    \item[Token Count] The total count of lexical tokens indicates the overall size of the code and its potential for human readability or interpretability.
    \item[Parameter Count] The number of parameters in functions and methods reflects the modularity and configurability of the code.
    \item[Function-Level Aggregates] Per-function averages and totals for cyclomatic complexity, token count, and parameter count are computed to understand variations within the codebase.
    \item[Depth and Nesting Metrics] The maximum and average levels of nesting provide insights into structural depth and potential readability challenges.
\end{description}

These complexity measures are computed with the Lizard~\cite{martin2017c} tool for static code analysis and further enriched by custom AST-based feature extraction. 
The total resulting code features we extract and use are listed in Table~\ref{tab:ASTfeatures}.

\begin{table*}[ht]
\centering
\caption{Features Extracted for Code Analysis\label{tab:ASTfeatures}}
\begin{tabular}{rp{12.5cm}}
\toprule
\textbf{Feature Name} & \textbf{Explanation} \\ \midrule
\textbf{Node Count} & Total number of nodes in the AST, reflecting the size of the code structure.\\ 
\textbf{Edge Count} & Total number of edges in the AST, representing the connections between syntactic elements.\\ 
\textbf{Edge Density} & Ratio of edges to nodes in the AST, providing a measure of graph connectivity.\\ 
\textbf{Min.|Max.|Mean|Var. Degree} & Statistical features about the degrees of the nodes in the AST.\\ 
\textbf{Degree Entropy} & Entropy of node degrees, capturing variability in connectivity.\\ 
\textbf{Assortativity} & Degree assortativity of the AST, measuring structural correlations between connected nodes.\\ 
\textbf{Min.|Max.|Mean. Depth} & Features regarding the depth of the AST, indicating the level of nested constructs.\\ 
\textbf{Depth Entropy} & Entropy of node depths, quantifying structural balance in the AST.\\ 
\textbf{Min.|Max.|Mean|Var. Clustering} & Clustering coefficient features of the AST nodes, indicating the modularity of code.\\ 
\textbf{Transitivity} & Global measure of clustering in the AST, representing code cohesion.\\ 
\textbf{Diameter} & Longest shortest path in the AST, representing the maximum distance between two nodes.\\ 
\textbf{Radius} & Shortest maximum distance from any node to all other nodes, indicating graph compactness.\\ 
\textbf{Mean Eccentricity} & Mean eccentricity of nodes (eccentricity is the longest distance from a node to any other node).\\ 
\textbf{Average Shortest Path} & Mean shortest path length in the AST, representing structural compactness.\\ 
\textbf{Cyclomatic Complexity} & Logical total and mean code complexity, representing the number of independent execution paths.\\ 
\textbf{Token Count} & Total and average number of tokens in the code, reflecting its size and verbosity.\\ 
\textbf{Parameter Count} & Number of parameters in functions and average number per function, reflecting configurability.\\ 
\bottomrule
\end{tabular}
\end{table*}

\subsection{Code Evolution Graphs}
\label{sec:code_evolution_graphs}

We define a \textit{Code Evolution Graph (CEG)} as a directed graph $G = (V, E)$, where:

\begin{itemize}
    \item $V$ is the set of nodes, each corresponding to an algorithmic instance represented by a vector of extracted features from its AST and associated metadata.
    The nodes $v_i \in V$ are defined by:
    \[
    v_i = (f_i, \mathbf{x}_i, m_i)
    \]
    where:
    \begin{itemize}
        \item $f_i \in [0, 1]$ is the normalized performance of the algorithm on the respective benchmark;
        \item $\mathbf{x}_i \in \mathbb{R}^d$ is the $d$-dimensional vector of standardized AST features, such as cyclomatic complexity, token count, parameter count, and other extracted metrics;
        \item $m_i$ is the \emph{metadata} associated with the algorithm, including identifiers (e.g., $\texttt{id}$ and $\texttt{name}$), evaluation index within the code optimization framework (such as LLaMEA) run, and parent IDs.
    \end{itemize}
    \item $E \subseteq V \times V$ is the set of directed edges, where an edge $e = (v_i, v_j)$ exists if algorithm $v_j$ was generated as a descendant of $v_i$ during an optimization process. Parent-child relationships are determined from the evolutionary lineage.
\end{itemize}



\textbf{Visualization:} To analyze the evolution of algorithms over the AAD evaluations, CEGs are projected into lower-dimensional spaces using dimensionality reduction techniques, in this case using Principal Component Analysis (PCA)~\cite{bro2014principal} and t-distributed Stochastic Neighbor Embedding (t-SNE)~\cite{cieslak2020t}. Nodes are plotted in the reduced space, with node sizes proportional to their frequency as parents in the evolutionary lineage.

\textbf{Purpose:} Code Evolution Graphs provide a structured representation of algorithm evolution, enabling the analysis of structural changes in the code, fitness improvements, and lineage over successive generations.

\section{Analysis of Different Code Evolution Strategies}

To evaluate the effectiveness of the Code Evolution Graphs and gain insights into algorithm design and evolution, we consider three benchmark tasks: Black-Box Optimization (BBO), Online Bin Packing (OBP), and the Traveling Salesperson Problem (TSP). Each benchmark represents a distinct optimization challenge, enabling a comprehensive analysis of the generated algorithms' adaptability, scalability, and structural complexity in the AAD setting.

For the visualizations in this work, we split the AST and complexity features and use the AST features as input for the PCA transformation. We choose to look at complexity separately as we saw a strong linear correlation of the different complexity features with each other. To analyse the complexity of the generated algorithms, we mainly look at code token counts, as the cyclomatic complexity and number of function parameters are highly influenced by the way the evolutionary frameworks are organized. For example, for the LLaMEA-HPO algorithm, the prompt asks explicitly to add any tunable parameter to the function prototype, causing a much higher number of parameters by construction.

\subsection{Black-Box Optimization}
\label{BBO}
The Black-Box Optimization benchmark suite consists of a set of continuous, noiseless functions that challenge the optimization capabilities of metaheuristic algorithms. The objective is to minimize a given function $f: \mathbb{R}^d \to \mathbb{R}$, where no information about the function’s structure or derivatives are available. Instead, algorithms must iteratively query the function and refine their search based solely on previously evaluated solutions. 

In the AAD task, the goal is to automatically generate and optimize metaheuristics capable of solving a diverse set of BBO functions. The algorithms evolved by frameworks such as LLaMEA and EoH are evaluated on their ability to balance exploration and exploitation across the black-box optimization landscape by evaluating them on diverse optimization problems. More specifically, the generated algorithms are evaluated on $24$ noiseless functions in a $5d$ search space using the BBOB suite \cite{bbob_hansen2009_noiseless} and using $3$ instances and $3$ independent runs for each of the benchmark functions (typical usage of the BBOB suite). Key challenges include navigating multimodal, separable, and highly-conditioned landscapes to achieve competitive performance within the evaluation budget (of $10\,000$ function evaluations). 
For additional details on the exact setup of this task, we refer the reader to \cite{vanstein2024llamealargelanguagemodel}.


\begin{figure}[t]
    \includegraphics[width=\linewidth,trim=0mm 0mm 00mm 0mm,clip]{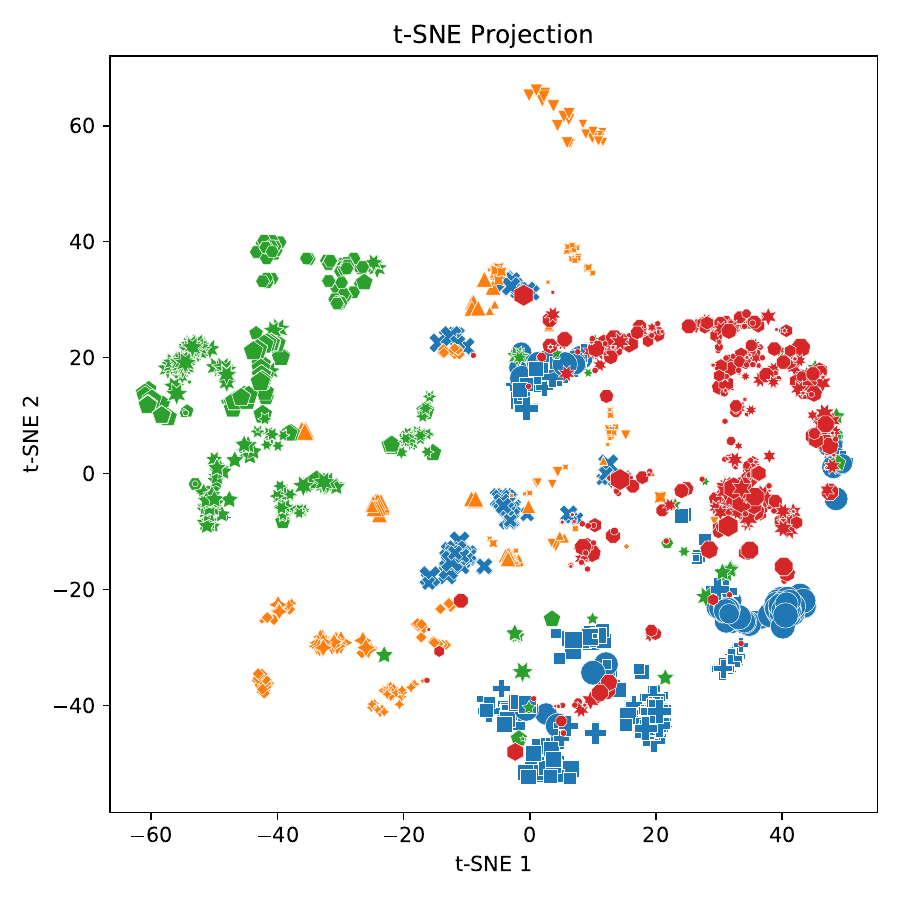}
    \includegraphics[width=.7\linewidth,trim=5mm 5mm 5mm 5mm,clip]{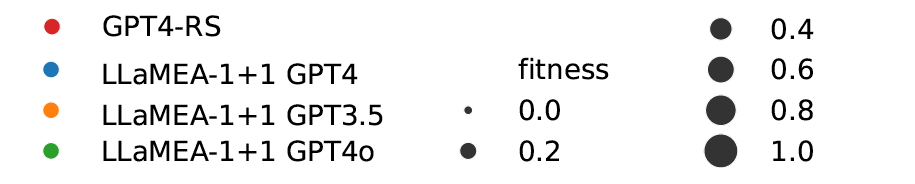}
    \caption{t-SNE visualisation of the $26$ code features for different LLaMEA configurations, $5$ independent runs per configuration and $5$ Random Search runs. The color denotes the method, different shapes denote different independent runs and different sizes denote different normalized fitness (bigger is better). \label{fig:tSNE}}
\end{figure}

Figure \ref{fig:tSNE} shows the t-SNE projection of the extracted $26$ code features for different LLaMEA configurations and Random Search (RS) runs. Each marker represents a single optimization algorithm, with different shapes denoting independent LLaMEA runs and marker sizes corresponding to fitness values (larger markers indicate higher fitness). The visualization reveals several insights into the behavior of the optimization process.

First, the runs are mostly non-overlapping, indicating distinct optimization trajectories for each configuration. Notably, only the RS experiments show some degree of overlap, suggesting that RS lacks the structured exploration seen in the LLaMEA-based methods.
Second, the solutions generated by different LLMs occupy distinct regions of the feature space, highlighting unique coding ``fingerprints'' characteristics of each LLM. This suggests that each LLM employs unique generation and refinement strategies, producing algorithmic solutions with different structural properties.
Third, the clustering of points within each optimization run suggest an ``evolutionary'' trend in the feature space. This pattern implies that the evolutionary loop and in-context learning mechanisms in LLaMEA effectively guide the search towards better solutions.
Finally, the diversity in coding fingerprints across LLMs supports the argument for leveraging multiple LLMs rather than relying on a single model. By utilizing multiple LLMs, one can exploit their complementary strengths, increasing the likelihood of generating diverse and high-quality solutions.



\begin{figure*}[t]
    \includegraphics[width=.48\textwidth,trim=0mm 0mm 0mm 0mm,clip]{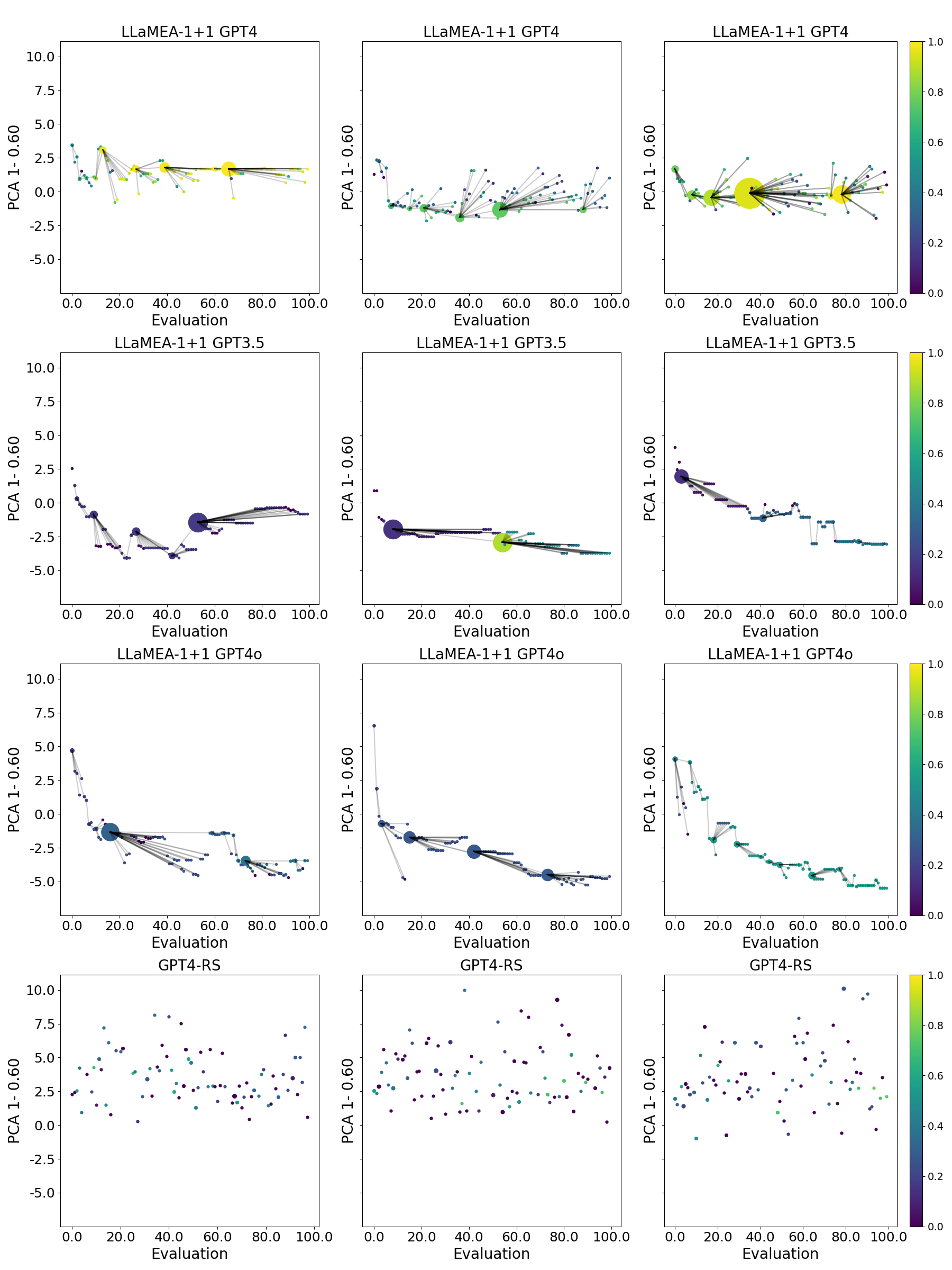}
    \includegraphics[width=.48\textwidth,trim=0mm 0mm 0mm 0mm,clip]{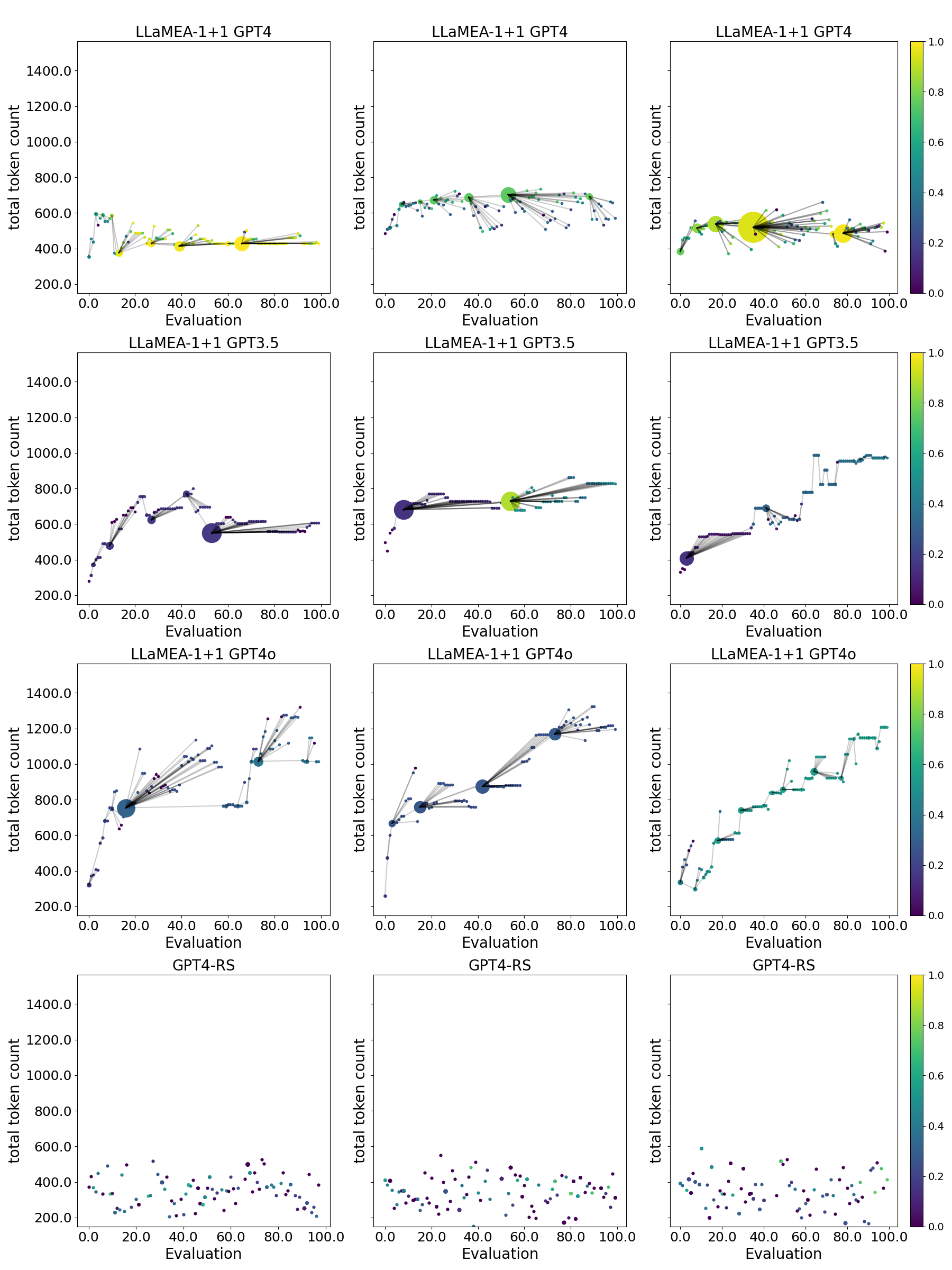}
    \caption{Code Evolution Graphs for LLaMEA using different LLMs and a baseline Random Search on BBO. On the left side are CEGs using the first PCA component of the AST graph metrics on the $y$-axis, with the number denoting the fraction of the total variance accounted for by this dimension. On the right side are CEGs using the total token count on the $y$-axis. Each row represents a different algorithm configuration (LLM) and each column is one independent run (3 runs in total). \label{fig:BBO}}
\end{figure*}


Figure~\ref{fig:BBO} presents the Code Evolution Graphs for the BBO benchmark. The left side visualizes the CEGs using the first principal component on the $22$ AST graph code features (without complexity features) as the $y$-axis, while the right side depicts the total token count on the $y$-axis to show an indicator of total code-complexity. The first PCA component coefficients can be found in our Zenodo repository \cite{anonymous_2025_14697614}, indicating which features explain most variability in the data.
Each row corresponds to a different AAD algorithm (from top to bottom; LLaMEA using GPT-4-Turbo, GPT-4o, GPT3.5 and Random Search using GPT-4-Turbo). Each column shows independent optimization runs. From these visualizations, we observe distinct structural patterns in the evolution of algorithms. LLaMEA uses by default a $1 + 1$ strategy with one parent generating one offspring algorithm. This results in a CEG where each node (algorithm) is connected to exactly one parent node. The graphs show how many mutations it took on a parent algorithm to finally improve the fitness. In the Random Search algorithm there is no notion of parents and offspring, and so they are not connected by edges.

\subsection{Online Bin Packing} 
\label{OBP}

The Online Bin Packing problem is a classic combinatorial optimization challenge where items of varying sizes arrive sequentially and must be placed into a finite set of bins with fixed capacity. The objective is to minimize the total number of bins used while adhering to capacity constraints. OBP requires algorithms to make decisions in real time, without knowledge of future items.

The AAD task for OBP involves generating heuristics that can efficiently determine bin placement strategies for incoming items. These heuristics are evaluated on their ability to generalize across multiple problem instances, where item sequences vary. The evolution process must identify effective scoring functions or decision rules that minimize bin usage while maintaining computational efficiency. For additional details on the exact setup of this task, we refer the reader to \cite{fei2024eoh}.


\begin{figure*}[t]
    \includegraphics[width=.48\textwidth,trim=0mm 0mm 0mm 0mm,clip]{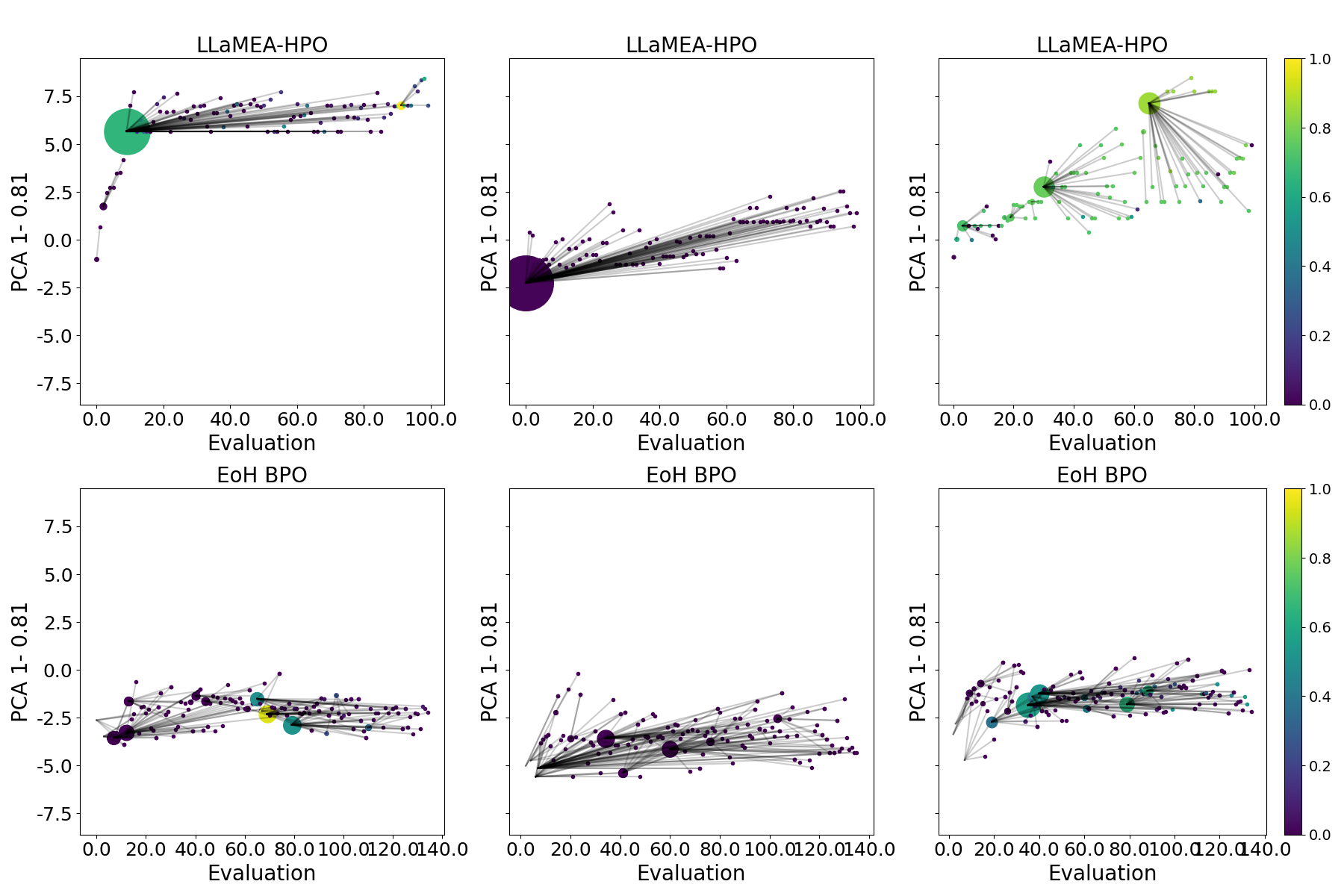}
    \includegraphics[width=.48\textwidth,trim=0mm 0mm 0mm 0mm,clip]{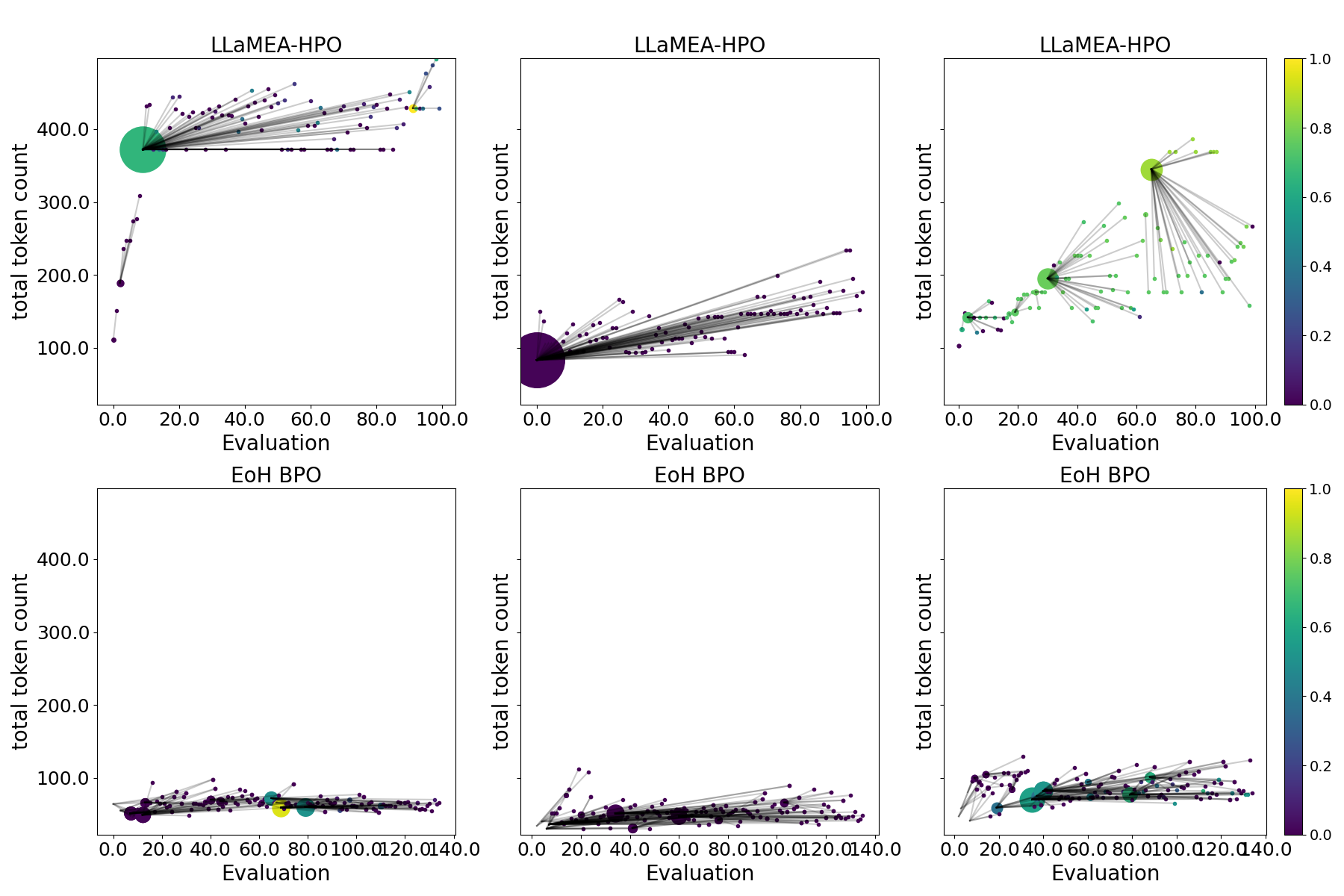}
    \caption{Code Evolution Graphs for LLaMEA-HPO and EoH on the Online Bin Packing Problems. On the left side are CEGs using the first PCA component of the AST graph metrics on the $y$-axis, with the number denoting the fraction of the total variance accounted for by this dimension. On the right side are CEGs using the total token count on the $y$-axis. The top row shows different runs of LLaMEA-HPO and the bottom row shows different runs for EoH. \label{fig:OBP}}
\end{figure*}



The Code Evolution Graphs for OBP are shown in Figure~\ref{fig:OBP}. Similar to the BBO visualizations, PCA projections and code token count analyses are used to capture the structural evolution of the algorithms. The CEGs illustrate how LLaMEA-HPO and EoH adapt their strategies over successive generations. 
Evolution of Heuristics uses by default a $4+20$ strategy with a population size of $4$ parents, and each parent undergoing $5$ different mutations to end up with $20$ offspring. This shows in the CEG as a much more densely connected graph. We can also observe that EoH generates mutations that are relatively close to each other and there is no clearly visible exploration in the PCA feature space compared to LLaMEA-HPO. Note that hyper-parameter optimization in LLaMEA-HPO only affects the fitness of the algorithms and not their features.

\subsection{Traveling Salesperson Problems}
\label{TSP}

The Traveling Salesperson Problem is a well-known NP-hard problem in which the objective is to determine the shortest possible route that visits a set of cities exactly once and returns to the starting point. Given a complete weighted graph, the task is to minimize the total length of the tour.

In the AAD setting, the challenge lies in generating heuristics that guide a local search algorithm, Guided Local Search (GLS), to explore the solution space effectively. The heuristics evolve strategies for dynamically penalizing frequently used edges, thereby encouraging exploration of alternative paths and improving convergence to near-optimal solutions. TSP serves as a benchmark for assessing the scalability and adaptability of generated algorithms across problem sizes and graph structures. For additional details on the exact setup of this task, we refer the reader to \cite{fei2024eoh}.


\begin{figure*}[t]
    \includegraphics[width=.48\textwidth,trim=0mm 0mm 0mm 0mm,clip]{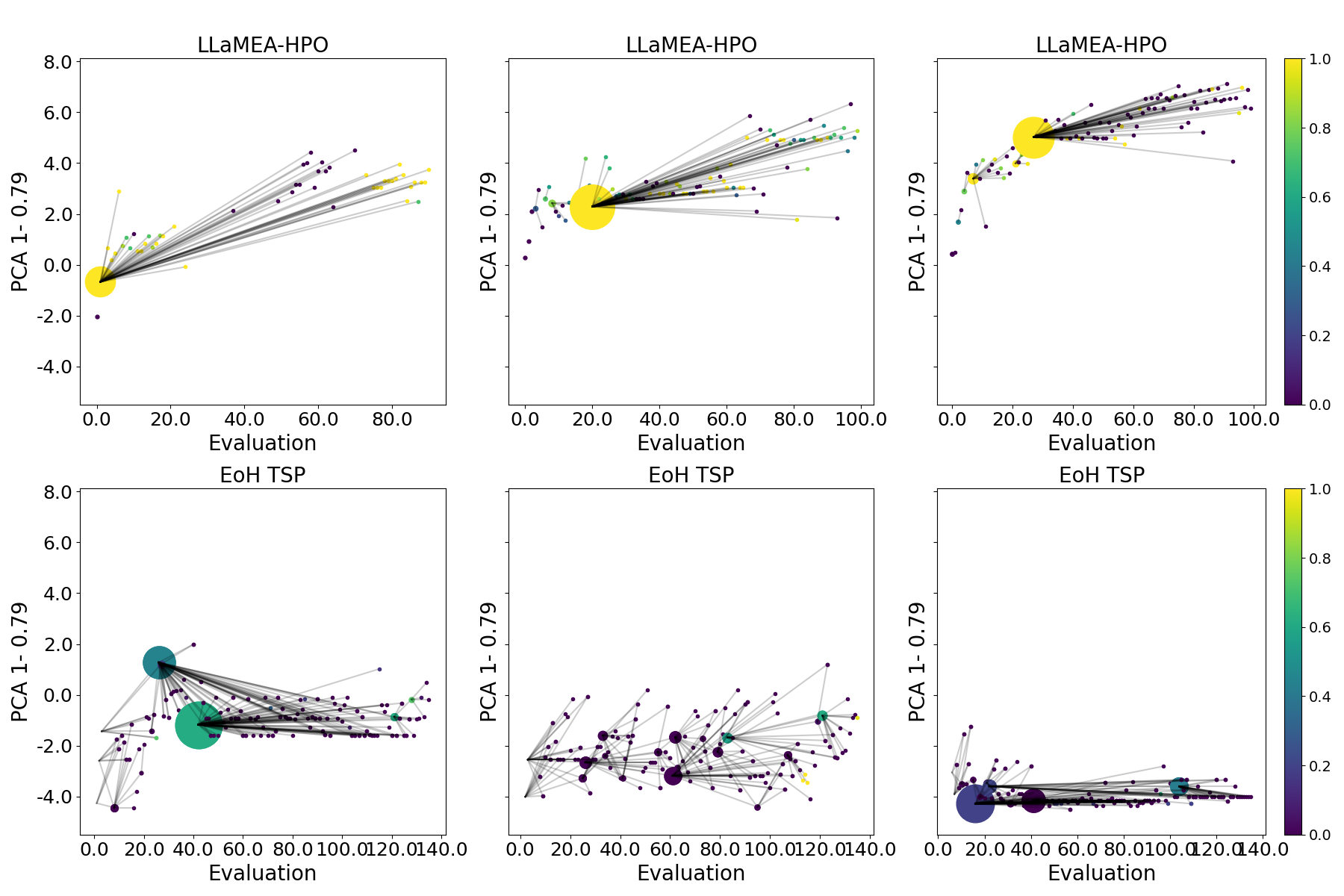}
    \includegraphics[width=.48\textwidth,trim=0mm 0mm 0mm 0mm,clip]{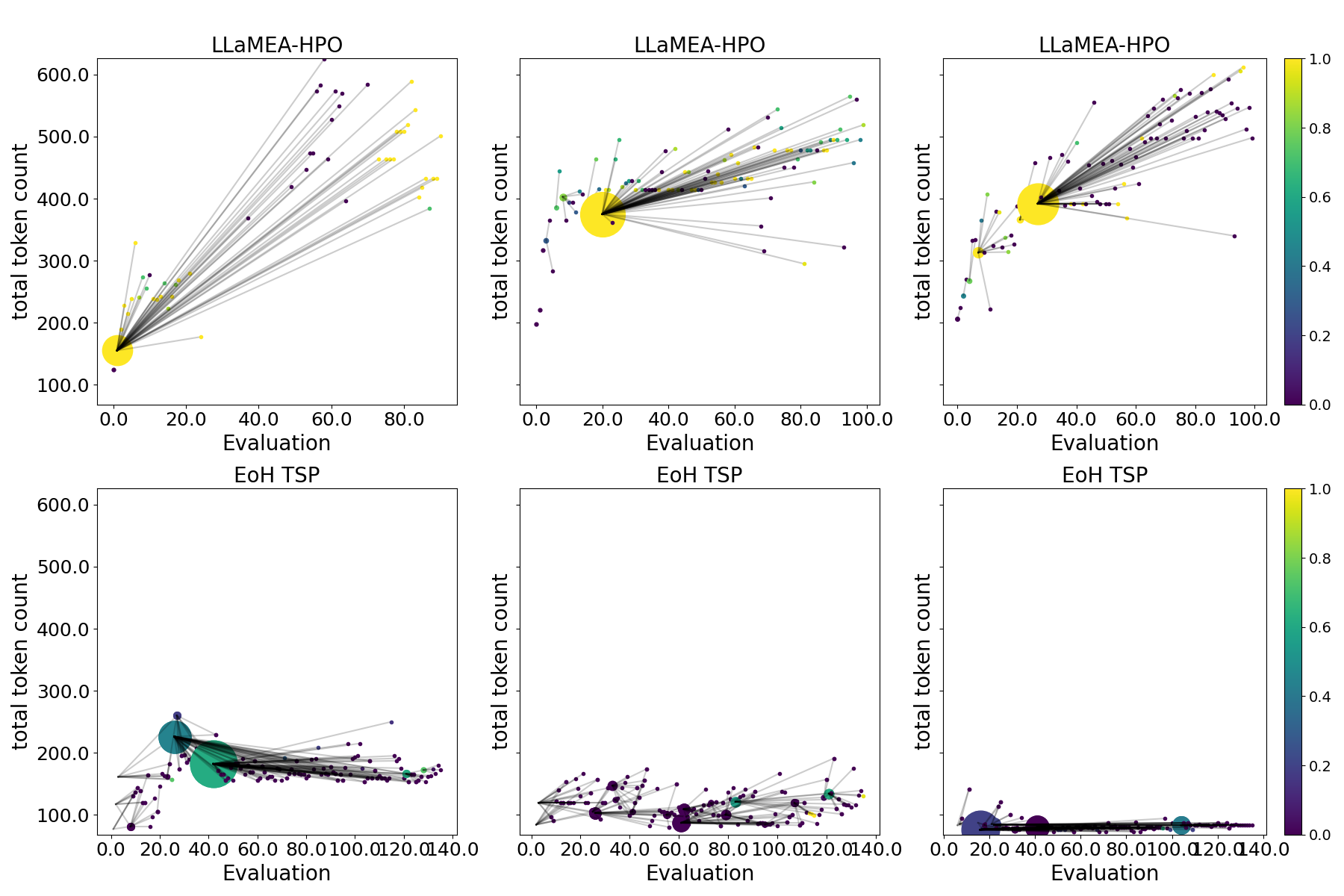}
    \caption{Code Evolution Graphs for LLaMEA-HPO and EoH on the Traveling Salesperson Problems. On the left side are CEGs using the first PCA component of the AST graph metrics on the $y$-axis, with the number denoting the fraction of the total variance accounted for by this dimension. On the right side are CEGs using the total token count on the $y$-axis. The top row shows different runs of LLaMEA-HPO and the bottom row shows different runs for EoH. \label{fig:TSP}}
\end{figure*}

Figure~\ref{fig:TSP} shows the Code Evolution Graphs for the TSP benchmark. Also in this case the EoH approach uses a population size of $4$ with $20$ offspring being generated each iteration. The CEGs from LLaMEA-HPO show that a near-optimal solution was found very early in the search process (in the first 20 evaluations) and that the search afterwards did not improve anymore. For EoH the search is less successful in finding a near-optimal solution very fast. We observe a clear trend of code complexity increasing with the number of evaluations.


\subsection{Analysis of Code Features}


\begin{figure*}[bt]
    \includegraphics[width=1\textwidth,trim=0mm 4mm 9cm 0mm,clip]{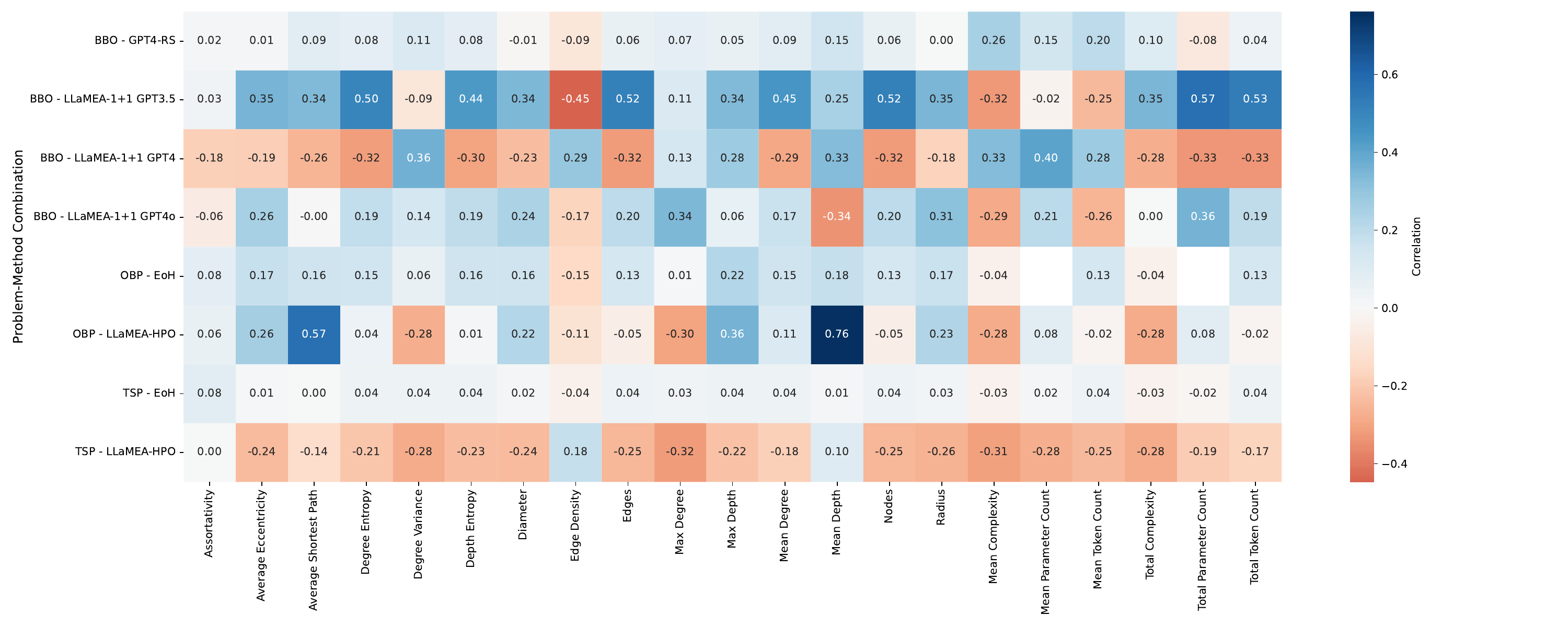}
   
    \caption{Spearman correlation index for each code feature (column) with the performance of the algorithms (fitness) for all benchmarks and methods (rows). \label{fig:correlation}}
\end{figure*}

To quantify how features affect the fitness of the generated algorithms, we compute the Spearman correlation between extracted code features and fitness values across all benchmark tasks and LLMs. The result is shown in Figure~\ref{fig:correlation}. In the BBO benchmark (top four rows), mean complexity, total parameter count, and mean parameter count emerge as the most influential, suggesting that algorithm modularity, size and logical complexity are key drivers of performance of the generated algorithms in continuous optimization tasks. The positive correlation for all those features suggests that more complex algorithms perform better. For the OBP benchmark (rows five and six), the structural properties max and mean depth play the most prominent roles, with a positive correlation suggesting again that more complex code tends to perform better. Finally, in the TSP benchmark (last two rows), max degree and mean complexity have a substantial impact. However, in contrast to the other two benchmarks, the correlation is negative, showing that less complex solutions for this benchmark provide better performance than more complex solutions.

The important code features and their impact vary across benchmarks and across LLMs. There is no single set of features that always affects performance in the same way. This makes sense, as we are solving very different benchmarks with starting codes with different performances. Again we observe that different LLMs have different ``fingerprints'' and tend to modify code in different ways. For example, the fitness of code for BBO LLaMEA $1+1$ GPT-4-Turbo is relatively strongly positively correlated with mean depth, while swapping GPT-4-Turbo for GPT-4o changes this to a relatively strong negative correlation. For TSP EoH, no feature shows a strong correlation with the fitness of the evolved algorithm, suggesting that changing the code does not affect performance much -- the LLM is unable to find changes that improve performance. In contrast, for BBO LLaMEA $1+1$ GPT-3.5 and GPT-4-Turbo in particular, many features have relatively strong positive or negative correlations with performance, suggesting that changes being made to the code have a large impact on performance.


\section{Conclusions and Future Work}

In this paper, we introduced Code Evolution Graphs as a novel methodology for analyzing and visualizing the structural evolution of algorithms generated by LLMs in evolutionary frameworks. By extracting and analyzing a comprehensive set of structural and complexity features from Abstract Syntax Trees, we were able to uncover insights into the design principles underlying successful algorithms for different tasks.

Our results demonstrated several findings:
\begin{itemize}
    \item \textbf{Diversity in Solutions:} CEGs reveal that the generated algorithms span a vast and non-overlapping feature space, emphasizing the diversity of code-space explored during the optimization process. This highlights the ability of frameworks such as LLaMEA and EoH to explore novel algorithmic structures.
    
    \item \textbf{Trends in Complexity:} We observed a consistent upward trend in code complexity (in particular, code token count) over the evolutionary process for LLaMEA-based frameworks, especially using the GPT-4o model. This suggests that the LLM-driven mutation strategies, particularly within the $1+1$ evolutionary paradigm, tend to produce increasingly complex algorithms over time.

    \item \textbf{Task-Specific Feature Importance:} Our analysis of the correlation of feature values with fitness showed that the importance of individual features varies across benchmarks. For the Black-Box Optimization and Online Bin Packing tasks, higher complexity correlated positively with fitness. In contrast, for the Traveling Salesperson Problem, simpler solutions with lower complexity performed better, suggesting that more complex algorithms are not always better.

    \item \textbf{Framework Comparisons:} LLaMEA-HPO, with its in-the-loop hyper-parameter optimization, exhibited smoother evolutionary trajectories, compared to EoH's reliance on structural diversity through mutations.
\end{itemize}

These insights demonstrate the utility of Code Evolution Graphs as a tool for linking the structure and complexity of algorithms to their fitness performance. By providing a systematic approach to analyzing code evolution, CEGs enable a better understanding of the strengths and limitations of LLM-based frameworks in the AAD domain, and actionable insights for improving AAD with LLMs.

While this work provides a foundation of analyzing LLM AAD frameworks, several avenues for future research remain:
\begin{itemize}
    \item \textbf{Code Features}: One of the main limitations of the representation by AST features is that a badly configured algorithm would have exactly the same AST features as a hyper-parameter optimized algorithm. Research into other (dynamic) features would strengthen the methodology.
    \item \textbf{Complexity Control:} Techniques to limit or regulate code complexity during the evolutionary process could improve the efficiency and interpretability of generated algorithms.
    \item \textbf{Diversity Management:} Strategies for controlling diversity in the population, such as diversity-aware mutation operators or clustering methods, could further enhance exploration behavior.
    \item \textbf{Informed LLM Mutations:} Incorporating feedback mechanisms to guide LLM-driven mutations towards more meaningful structural changes.
\end{itemize}

This work bridges the gap between algorithmic evolution and code structure analysis, offering a  framework for visualizing and analyzing LLM-driven algorithm design tasks.
All code and additional results are available in our Zenodo repository \cite{anonymous_2025_14697614}.



\begin{acks}
This work has benefitted from Dagstuhl Seminar 24282, ``Automated Machine Learning for Computational Mechanics''. We thank Grace Abawe for writing the original AST analysis code.
\end{acks}

\FloatBarrier

\bibliographystyle{ACM-Reference-Format}
\bibliography{refs}

\end{document}